\documentclass[10pt,twocolumn,letterpaper]{article}

\usepackage{cvpr}
\usepackage{times}
\usepackage{booktabs} 

\usepackage{url}
\usepackage{courier}
\usepackage{graphicx}
\usepackage{epstopdf}
\usepackage[titletoc]{appendix}
\usepackage{epsfig}
\usepackage{picinpar}
\usepackage{amsmath}
\usepackage{amsthm}
\usepackage{amssymb,bm}
\usepackage[ruled,vlined]{algorithm2e}
\usepackage[titletoc]{appendix}
\usepackage{float,caption,multirow}
\usepackage{subcaption,xcolor}

\newcommand{\defeq}{\mathrel{\mathop:}=}

\newcommand{\mc}{\mathcal}
\newcommand{\mb}{\mathbb}

\cvprfinalcopy 

\def\cvprPaperID{351} 

\ifcvprfinal\fi
\begin{document}

\title{EDIT: Exemplar-Domain Aware Image-to-Image Translation}

\author{Yuanbin Fu\\
Tianjin University\\
{\tt\small fu199671@126.com}
\and
Jiayi Ma\\
Wuhan University\\
{\tt\small jyma2010@gmail.com}
\and
Lin Ma\\
Tencent AI Lab\\
{\tt\small forestlma@tencent.com}
\and
Xiaojie Guo\\
Tianjin University\\
{\tt\small  xj.max.guo@gmail.com}
}

\maketitle

\begin{abstract}
Image-to-image translation is to convert an image of the certain style to another of the target style with the content preserved. A desired translator should be capable to generate diverse results in a controllable (many-to-many) fashion. To this end, we design a novel generative adversarial network, namely exemplar-domain aware image-to-image translator (EDIT for short). The principle behind is that, for images from multiple domains, the content features can be obtained by a uniform extractor, while (re-)stylization is achieved by mapping the extracted features specifically to different purposes (domains and exemplars). The generator of our EDIT comprises of a part of blocks configured by shared parameters, and the rest by varied parameters exported by an exemplar-domain aware parameter network. In addition, a discriminator is equipped during the training phase to guarantee the output satisfying the distribution of the target domain. Our EDIT can flexibly and effectively work on multiple domains and arbitrary exemplars in a unified neat model.  We conduct experiments to show the efficacy of our design, and reveal its advances over other state-of-the-art methods both quantitatively and qualitatively.
\end{abstract}

\section{Introduction}

A scene can be expressed in various manners using sketches, semantic maps, photographs, and painting artworks, to name just a few. Basically, the way that one portrays the scene and expresses his/her vision is so-called style, which can reflect the characteristic of either a class/domain or a specific case. Image-to-image translation (I2IT) \cite{NIPS2018_7404}\cite{NIPS2018_7627}\cite{alharbi2019latent}\cite{wu2019relgan}\cite{Shen_2019_CVPR} refers to the process of converting an image $I$ of the certain style to another of the target style $S_t$ with the content preserved. Formally, seeking a desired translator  ${\mc{T}}$ can be written in the following form:
\begin{equation}
\begin{aligned}
\min~ \mc{C}(I_t, I)+\mc{S}(I_t, S_t)
~~\text{with}~~I_t\defeq\mc{T}(I, S_t),
\end{aligned}
\label{eq:define}
\end{equation}  where $\mc{C}(I_t, I)$ is to measure the content difference between the translated $I_t$ and the original $I$, while $\mc{S}(I_t, S_t)$ is to enforce the style of  $I_t$ following that indicated by $S_t$.

\subsection{Previous Arts}

\begin{figure*}[t]
	\begin{center}
		\centering
		\includegraphics[width=\textwidth]{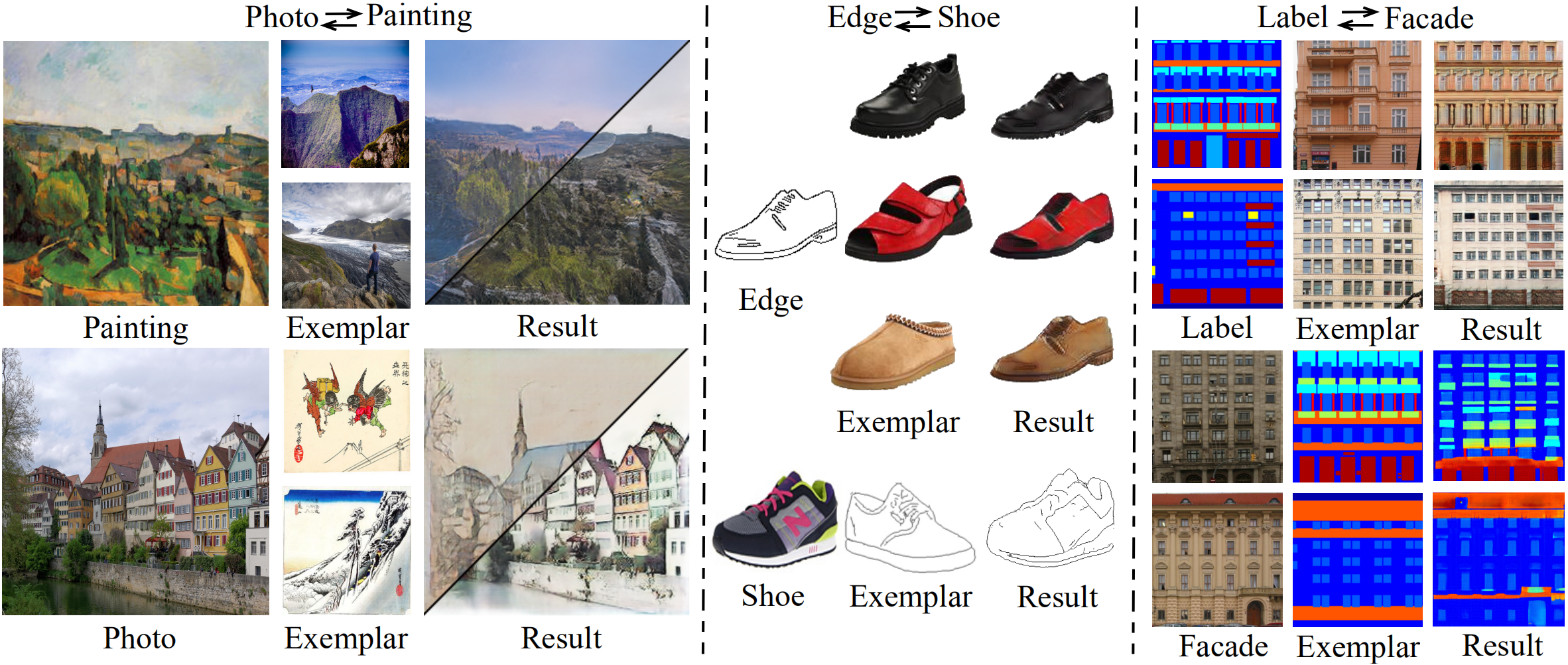}
		\caption{Several results by the proposed EDIT. Our EDIT is able to take arbitrary exemplars as reference for translating images across multiple domains including photo-painting, shoe-edge, and semantic map-facade in \emph{one} model.}
		\vspace{-10pt}
		\label{fig:teaser}
	\end{center}
\end{figure*}

With the emergence of deep techniques, a variety of I2IT strategies have been proposed with great progress made over the last decade. In what follows, we briefly review contemporary works along two main technical lines, \textit{i.e.} one-to-one translation and many-to-many translation.

\textit{One-to-one Translation.} Methods in this category aim at mapping images from a source {domain} to a target {domain}. Benefiting from the generative adversarial networks (GANs) \cite{goodfellow2014generative}, the style of translated results satisfies the distribution of the target domain $\mb{Y}$, achieved by $\mc{S}(I_t, S_t)\defeq\mc{D}(I_t, \mb{Y})$, where $\mc{D}(I_t, \mb{Y})$ represents a discriminator to distinguish if $I_t$ is real with respect to  $\mb{Y}$. An early attempt by Isola \textit{et al.} \cite{isola2017image} uses conditional GANs to learn mappings between two domains. The content preservation is supervised by the paired data, \textit{i.e.} $\mc{C}(I_t, I)\defeq\mc{C}(I_t, I_t^{gt})$ with $I_t^{gt}$ the ground-truth target. However, in real-world situations, acquiring such paired datasets, if not impossible, is impractical. To alleviate the pressure from data, inspired by the concept of cycle consistency, cycleGAN \cite{zhu2017unpaired} in an unsupervised fashion was proposed, which adopts $\mc{C}(I_t, I)\defeq\mc{C}(\mc{F}_{\mb{Y}\rightarrow\mb{X}}(\mc{F}_{\mb{X}\rightarrow\mb{Y}}(I)),I)$ with $\mc{F}_{\mb{X}\rightarrow\mb{Y}}$ the generator from domain $\mb{X}$ to $\mb{Y}$ and $\mc{F}_{\mb{Y}\rightarrow\mb{X}}$ the reverse one. Afterwards, StarGAN \cite{choi2018stargan} further extends the translation between two domains to that cross multiple domains in a single model. Though the effectiveness of the mentioned methods has been witnessed by a wide spectrum of specific applications such as photo-caricature \cite{cao2018carigans,inproceedings}, making up-makeup removal \cite{chang2018pairedcyclegan}, and face manipulation \cite{Wang_2018_CVPR}, {their main drawback comes from the nature of deterministic (uncontrollable) one-to-one mapping}.

\textit{Many-to-many Translation.} The goal of approaches in this class is to transfer the style controlled by an exemplar image to a source image with content maintained. Arguably, the most representative work goes to \cite{gatys2016image}, which uses the pre-trained VGG16 network \cite{Simonyan2014Very} to extract the content and style features, then transfer style information by minimizing the distance between Gram matrices constructed from the generated image and the exemplar $E$, say $\mc{S}(I_t, S_t)\defeq\mc{S}(\text{Gram}(I_t), \text{Gram}(E))$. Since then, numerous applications on 3D scene \cite{chen2018stereoscopic}, face swap \cite{korshunova2017fast}, portrait stylization \cite{Shih2014Style} and font design \cite{azadi2018multi} have been done. Furthermore, a number of schemes have also been developed towards relieving the limitations of \cite{gatys2016image} in terms of speed and flexibility. For example, to tackle the requirement of training for every new exemplar (style),  Shen \textit{et al.} \cite{shen2017meta} built a meta network, which takes in the style image and produces a corresponding image transformation network directly. Risser \textit{et al.} \cite{Risser2017Stable} proposed  the histogram loss to improve the training instability. Huang and Belongie \cite{AdaIN} designed a more suitable normalization manner, \textit{i.e.} AdaIN, for style transfer.  Li \textit{et al.} \cite{li2017demystifying} replaced the Gram matrices with an alternative distribution alignment manner from the perspective of domain adaption. Johnson \textit{et al.} \cite{johnson2016perceptual} trained the network with a specific style image and multiple content images while keeping the parameters at the inference stage. Chen \textit{et al.} \cite{Chen2017StyleBank} introduced a style-bank layer containing several filter-banks, each of which represents a specific style. Gu \textit{et al.} \cite{Gu2018Arbitrary} proposed a style loss to make parameterized and non-parameterized methods complement to each other. Huang \textit{et al.} \cite{Huang2017Real} designed a new temporal loss to ensure the style consistency between frames of a video. In addition, to mitigate the deterministic nature of one-to-one translation, several works, for instance  \cite{lee2018diverse}, \cite{lin2018conditional} and \cite{huang2018multimodal}, advocated to separately take care of content  $c(I)$ and style $s(I)$ subject to $I\simeq c(I)\circ s(I)$ with $\circ$ the combination operation. They manage to control the translated results by combining the content of an image with the style of target, \textit{i.e.} ${c}(I)\circ{s}(E)$. {Besides one domain pair requires one independent model, their performance, as observed from comparisons, is inferior to our method  in visual quality, diversity, and style preservation.} Please see Fig. \ref{fig:teaser} for examples produced by our method that handles multiple domains and arbitrary exemplars in a unified model.

\subsection{Challenges \& Motivations}

Developing a practical I2I translator remains challenging, because the capabilities of preserving content information after translation, and handling multiple domains as well as arbitrary exemplars should be considered jointly. We list the challenges as follows:
\begin{itemize}
	\item How to rationally disentangle the content and style representations of images from different domains in a unified fashion (multi-domain in one model)?
	\item How to effectively ensure the content of the translated result being consistent with that of the original image in an unsupervised manner (content preservation)?
	\item How to flexibly manipulate an image by considering both the style of a target domain and that of a specific exemplar (exemplar-domain style awareness)?
\end{itemize}

\begin{figure*}[t]
	\begin{center}
		\includegraphics[width=1\linewidth]{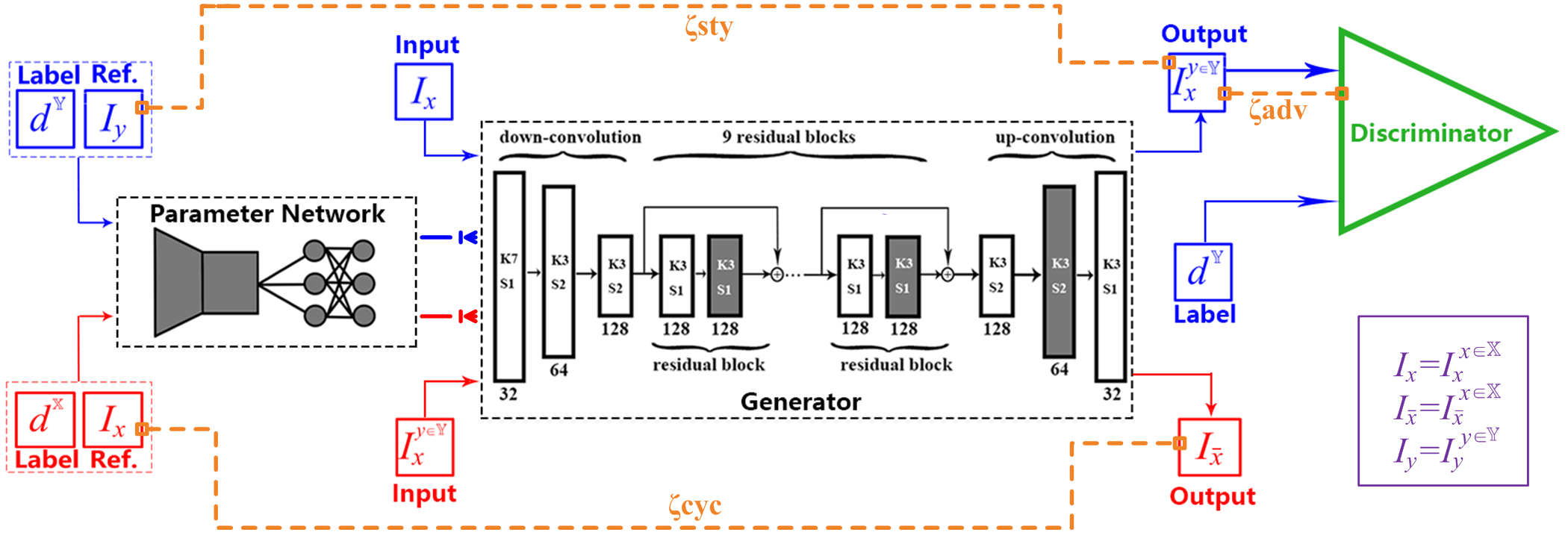}
	\end{center}
	\caption{The model architecture of our EDIT. The procedure of mapping $\mb{X} \rightarrow \mb{Y}$ is in blue, while the reverse of mapping $\mb{Y} \rightarrow \mb{X}$ is in red. $I_x$ and $I_y$ are samples from domain $\mb{X}$ and $\mb{Y}$, respectively. The whole network comprises of a generator and a discriminator. The generator contains a part of blocks configured by shared parameters, and the rest by varied parameters exported by an exemplar-domain aware parameter network. The parameter network generates the specific parameters based on an exemplar and its domain label. The content is preserved by adopting the cycle consistency. The discriminator takes a generated result and its domain label as input to judge if the result is distinguishable from the target domain. $K_k$ means that the kernel size is $k\times k$, while $S_s$ represents that the stride is $s$. The number of channels is given below each block.}
	\label{newpipline}
\end{figure*}

 Our principle is that, for images from different domains, the content features can be obtained by a uniform extractor, while (re-)stylization is achieved by mapping the extracted features specifically to different purposes. This principle is rational: taking artwork composition for an example, given a fixed scene, the physical content is the same, but the styles of presentation can be much diverse by different artists. For the style factor, one may generally like the paintings by Monet (domain), and among so many pieces of art, a particular one, \textit{e.g.} ``Water Lilies" (exemplar), is his/her favorite. In other words, the domain-level and exemplar-level should be simultaneously concerned during style transfer.  Moreover, to maintain the content information after translation, the cycle consistency can be employed due to its effectiveness and simplicity. It is worth emphasizing that, a single generator instead of a pair, like cycleGAN \cite{zhu2017unpaired}, could be sufficient if the content and style are well-disentangled.  

\subsection{Contributions}
Motivated by the above principle, we propose a novel network to overcome the mentioned challenges. Concretely, our primary contributions can be summarized as follows:
\begin{itemize}
	\item We design a network, namely EDIT, to produce diverse results in an unsupervised controllable (many-to-many) fashion, which can flexibly and effectively work on multiple domains and arbitrary exemplars in a unified model.
	\item The generator of our EDIT comprises of a part of blocks configured by shared parameters to uniformly extract features for images from multiple domains, and the rest by varied parameters exported by an exemplar-domain aware parameter network to catch specific style information.
	\item To preserve the content between input and generated result, the cycle consistency is employed. Plus, a discriminator is equipped during the training phase to guarantee the output satisfying the distribution of the target domain.
	\item We conduct extensive experimental results to reveal the efficacy of our design, and demonstrate its advantages over other state-of-the-art methods both quantitatively and qualitatively.
\end{itemize}
Several previous works, with feature transform in \cite{WCT}, style decoration in \cite{Avatar}, and feature normalization transfer in \cite{SCST} as representatives, insert an extra step, say feature manipulation, \textbf{between} the trained encoder and decoder to achieve style transfer, the spirit of which is seemingly similar but much different to ours. We achieve the style transfer \textbf{in} the decoder dynamically generated.
Even if there is no exemplar provided, the model still can produce results according to the target domain (by setting the exemplar to \textit{e.g.} a black image), which is more flexible than traditional style transfer methods like \cite{WCT,Avatar} having no domain information considered. Please notice that although the method \cite{SCST} considers both the domain and exemplar knowledge, it requires to train different models for different domain pairs, while our method is able to embrace multiple domain pairs in one neat model.

\section{Methodology}

\subsection{Problem Analysis}
\begin{figure}[t]
	\begin{center}
		\includegraphics[width=1\linewidth]{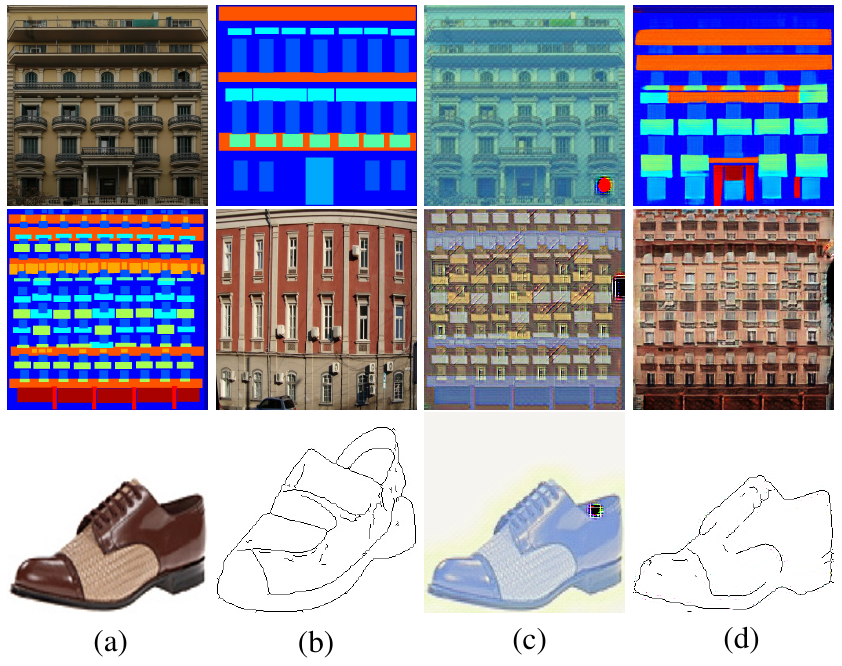}
	\end{center}
	\vspace{-7pt}
	\caption{Visual results by EDIT with and without specifying the target domain. (a) and (b) contain the inputs $I_x$ and exemplars $I_y$, respectively. (c) and (g) give the translated results without and with domain specification, respectively.}
	\vspace{-7pt}
	\label{fig:wod}
\end{figure}

A desired translator should be capable to generate diverse results in a controllable (many-to-many) fashion. Again, we emphasize the core principle behind this work: \textit{for images from different domains, the content features can be obtained by a uniform extractor, while (re-)stylization is achieved by mapping the extracted features specifically to different purposes}. In other words, we assume that the content $c(\cdot)$ and the style $s(\cdot)$ of an image are independent, \textit{i.e.} $p(I)=p(c(I), s(I))=p(c(I))\cdot p(s(I))$. Suppose that the whole style space is $\bigcup_i\mb{S}_i$, where $\mb{S}_i$ is the style subspace corresponding to the domain $i$. Mathematically, the problem can be expressed and solved by maximizing the following probability:
\begin{equation}
\begin{aligned}
p(I_x^y&|I_x,I_y)\defeq p(c(I_x^y), s(I_x^y)|c(I_x),s(I_x),c(I_y),s(I_y))\\
&\propto p(c(I_x^y)|c(I_x))\cdot p(s(I_x^y)|s(I_y))\\
&=p(c(I_x^y)|c(I_x))\cdot\sum_{i} p(s(I_x^y)|s(I_y), \mb{S}_i)\cdot p(\mb{S}_i)\\
&=\sum_{i} p(c(I_x^y)|c(I_x))\cdot p(s(I_x^y)|s(I_y), \mb{S}_i)\cdot p(\mb{S}_i).
\end{aligned}
\label{eq:core}
\end{equation}
The relationship of second row holds by the problem definition in Eq. \eqref{eq:define} and the independence assumption (our core principle). 
Furthermore, the style of $I_y$ may appear in more than one domains, for instance, a semantic map can also be a painting. This situation makes $ p(s(I_x^y)|s(I_y))$ a mixture of $\sum_i p(s(I_x^y)|s(I_y), \mb{S}_i)\cdot p(\mb{S}_i)$ (the equality of $3$rd row). Please see Figure \ref{fig:wod} for evidence. Therefore, we specify the domain label to clear the mix-up. By doing so, the problem turns to maximize the following:
\begin{equation}
\begin{aligned}
p(I_x^{y\in\mb{S}_i}|I_x, I_y&\in\mb{S}_i)\\
\defeq&
p(c(I_x^y)|c(I_x))\cdot p(s(I_x^y)|s(I_y), \mb{S}_i).
\end{aligned}
\label{eq:core2}
\end{equation}
As given in Eq. \eqref{eq:core2}, the entire problem can thus be divided into two subproblems. The first component $ p(c(I_x^y)|c(I_x))$ corresponds to the uniform content extractor, while the second term $ p(s(I_x^y)|s(I_y), \mb{S}_i)$ yields the exemplar-domain aware style mapping. 

\subsection{Architecture Design}
The blueprint of our EDIT is schematically illustrated in Figure \ref{newpipline}, from which, we can see that the generator of EDIT $\mc{G}$ is composed by a part of blocks configured by shared parameters $\theta_s$, and the rest by varied parameters $\theta_p$ exported by an exemplar-domain aware parameter auxiliary network.  In addition, a discriminator $\mc{D}$ is equipped during the training phase to guarantee the output satisfying the distribution of the target domain. 

The  \textbf{generator} is to produce desired images through
\begin{equation}
I_x^{y\in\mb{S}_i}\defeq\mc{G}(I_x, I_y\in\mb{S}_i; \theta),
\end{equation}
where $\theta$ is the trainable parameters for the whole generator. The generator consists of three gradually down-sampled encoding blocks, followed by $9$ residual blocks. Then, the decoder processes the feature maps gradually up-sampled to the same size as input. Each block performs in the manner of \textit{Conv+InstanceNorm+ReLU}. As stated, a part of the generator should respond to extract features uniformly for images no matter what styles they are in. In other words, a number of blocks (in white as shown in Fig. \ref{newpipline},  \textbf{uniform content extractor}) are shared across domains, the parameter set of which is denoted by $\theta_s$.  As for the rest blocks (in black as shown in Fig. \ref{newpipline}) related to (re)-stylization (feature selection and reassembling). Inspired by \cite{Hyper,shen2017meta}, the corresponding parameters can be dynamically generated by a parameter network, that is:
\begin{equation}
\theta_p\defeq\mc{G}_{P}(I_y\in\mb{S}_i;\psi),
\end{equation}
where $\psi$ is its trainable parameters. Please notice that, our parameter network only covers a part, instead of all as \cite{Hyper,shen2017meta}, of blocks in the generator, which significantly saves the resource. Specifically, the \textbf{parameter network} contains the \textit{VGG16} network  pre-trained on the ImageNet and fixed, followed by \textit{one fully connected layer and one group fully connected layer}. Feeding an exemplar (style image) and its target domain label (a one-hot vector) into the parameter network gives the parameters required by the \textbf{exemplar-domain aware style mapping}. Now, we can express the generator as follows:
\begin{equation}
I_x^{y\in\mb{S}_i}\defeq\mc{G}(I_x; \theta_p\defeq\mc{G}_{P}(I_y\in\mb{S}_i;\psi), \theta_s),
\end{equation}
where both $\theta_s$ and $\theta_p$ form $\theta$.

Based on the analysis on domain specification, it is important to clear the style mix-up issue as revealed in Fig. \ref{fig:wod}. Merely providing the domain ID to the parameter network is insufficient to capture the domain characteristic, as it is blind to the distribution of the target domain. To guide the training process and produce high-quality images satisfying the distribution of the target domain, we further employ a \textbf{discriminator} built upon the  $70\times70$ Patch-GAN architecture \cite{isola2017image}, which tries to determine whether each local image patch , rather than the whole image, is real or fake. More details about the discriminator can be found in the corresponding paper or at our website\renewcommand{\thefootnote}{\fnsymbol{footnote}}\textsuperscript{\ref{web}}. It is worth noting that the exemplar-domain aware style mapping is actually achieved by the dynamic part in the generator together with the discriminator.

One may wonder why inserting dynamic (black) blocks into fixed (white) blocks. First, considering the generation of dynamic parameters, the complexity of the fully connected layers will dramatically grow as the number of dynamic parameters (\textit{e.g.} all the blocks in the decoder) required to generate increases. From another point of view, the style mapping can be viewed as a procedure of feature selection and reassembling. Some operations should be in common for features from different domains. Taking the above concerns, we adopt the organization fashion as shown in Fig. \ref{newpipline}, which performs sufficiently well in practice and makes the volume of the parameter network compact. The primary merit of our EDIT is that it can handle arbitrary exemplars and be trained for multiple domains at the same time in \emph{one} neat model. More details of EDIT are given in supplementary.

\subsection{Loss Design}
We adopt a combination of a cycle consistency loss, a style loss and an adversarial loss for training the network.


\noindent\textbf{Cycle consistency loss.} Taking a sample pair $I_x\in\mb{X}$ and $I_y\in\mb{Y}$ for an example, let $I_{\bar{x}}$ and $I_{\bar{y}}$ be $\mc{G}(\mc{G}(I_x,I_y\in\mb{Y}),I_x\in\mb{X})$ and $\mc{G}(\mc{G}(I_y,I_x\in\mb{X}),I_y\in\mb{Y})$, respectively. To preserve content between generated and original images, the cycle consistency loss is employed,  which is written as:
\begin{equation}
\begin{aligned}
\zeta_{cyc}\defeq \| I_{\bar{x}} -I_x  \|_1 +\| I_{\bar{y}} - I_y \|_1,
\end{aligned}
\label{cyc}
\end{equation}
where $\|\cdot\|_1$ is the $\ell_1$ norm.


\begin{figure*}[t]
	\centering
	\centering
	\includegraphics[width=1\linewidth]{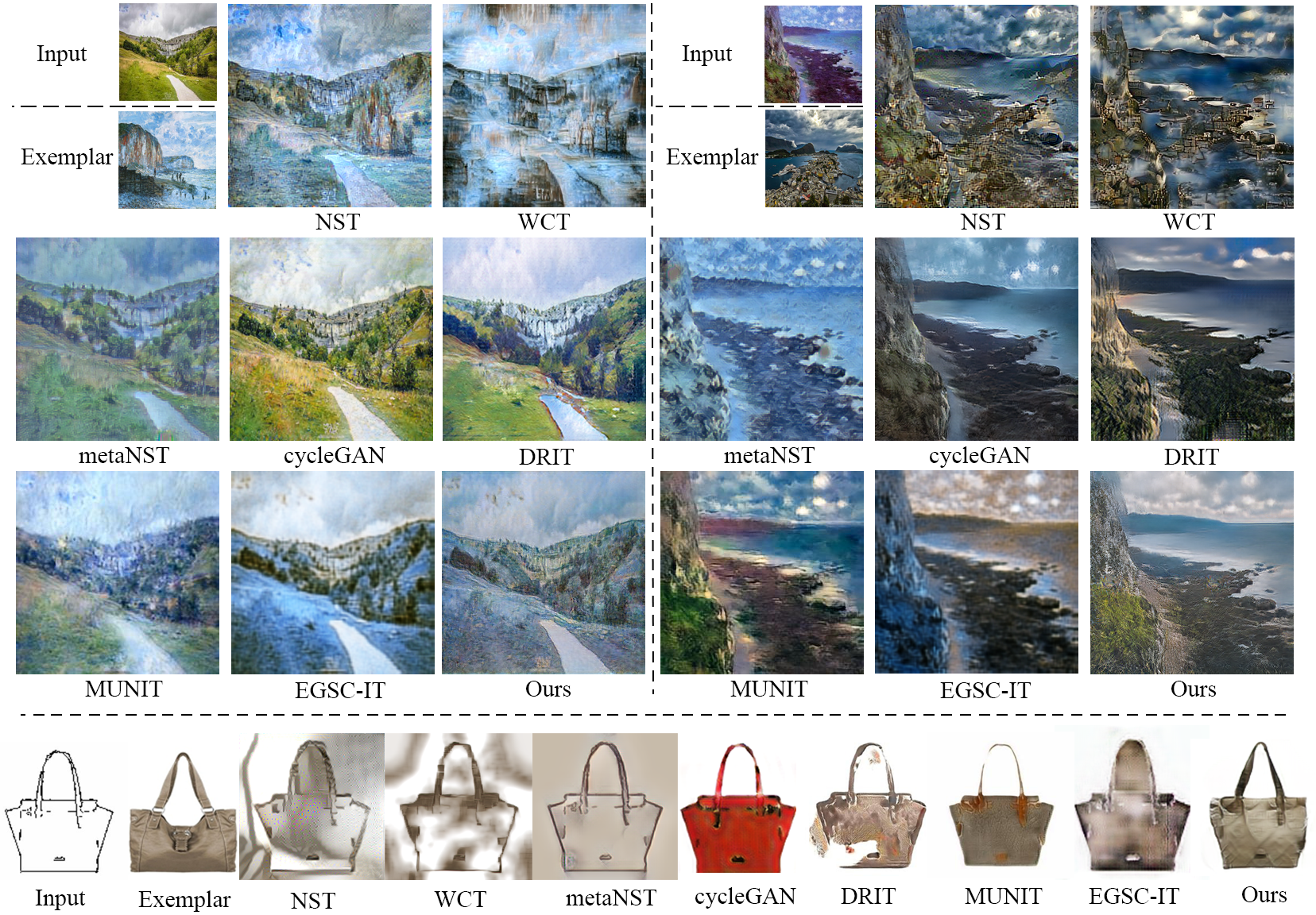}
	\caption{Visual comparison among the competitors on photo to painting, painting to photo, and edge to handbag.}
	\label{fig:vcomp}
\end{figure*}
\begin{table*}[t]
	\centering
	\begin{tabular}{l||c|cc|ccc}
		\hline
		Methods & Time (sec.) & Paras. (MB$|$1 pair)&  ($n=4$ pairs)& Content Error $\downarrow$ & Style Error $\downarrow$ & IncepScore $\uparrow$\\
		\hline
		\hline
		art2real \cite{Tomei_2019_CVPR}& $1.2\times10^{-2}$ & 45.5 & $-$&$-$ &$-$ & $-$\\
		cycleGAN \cite{zhu2017unpaired}& $3.5\times10^{-3}$ & 45+45& (45+45)$\times n$& 1.70$\pm$0.60  & $-$& 6.02\\
		MUNIT \cite{huang2018multimodal}& $3.9\times10^{-2}$ & 114.7& 114.7$\times n$& 2.43$\pm$1.28 & 0.19$\pm$0.15 & 4.58\\
		DRIT \cite{lee2018diverse}& $1.2\times10^{-2}$ & 780 & 780$\times n$& 2.83$\pm$1.17 & 0.14$\pm$0.09 & 5.06\\
		NST \cite{gatys2016image} & $4.3\times10^{2}$ & 576 & 576$\times n$ & 3.43$\pm$1.04 & 1.28$\pm$0.56 & 5.85\\
		metaNST \cite{shen2017meta}& $5.3\times10^{-3}$ & 64K+10[+870] & 64K+10[+870] & 2.97$\pm$0.93 & 0.13$\pm$0.09 & 4.74\\
		WCT \cite{WCT}& $1.7\times10^{0}$ & 283.6& 283.6 & 4.92$\pm$0.15 & 0.13$\pm$0.07 & 3.09\\
		EGSC-IT \cite{SCST} & $8.2\times10^{-1}$ & 135 & 135$\times n$& 2.71$\pm$1.29 & 0.26$\pm$0.19 & 5.50\\
		\hline
		\hline
		EDIT w/o Adv & $4.3\times10^{-3}$&8+3.6[+476] & 8+3.6[+476] & 3.39$\pm$0.98& 0.26$\pm$0.10 & 4.59\\
		\textbf{EDIT} & $4.3\times10^{-3}$ & 8+3.6[+476]& 8+3.6[+476]& {2.33$\pm$0.98} & {0.09$\pm$0.07} & {5.90}\\
		\hline
	\end{tabular}
	\caption{Quantitative comparison with the state-of-the-art methods. The two columns of model size (parameters) are for one domain pair and $n$ domain pairs, respectively. For the content and style errors, lower values indicate better performance. While for the inception score, the higher the better.}
	\label{tab:comp}
\end{table*}

\begin{figure*}[t]
	\centering
	\includegraphics[width=1\linewidth]{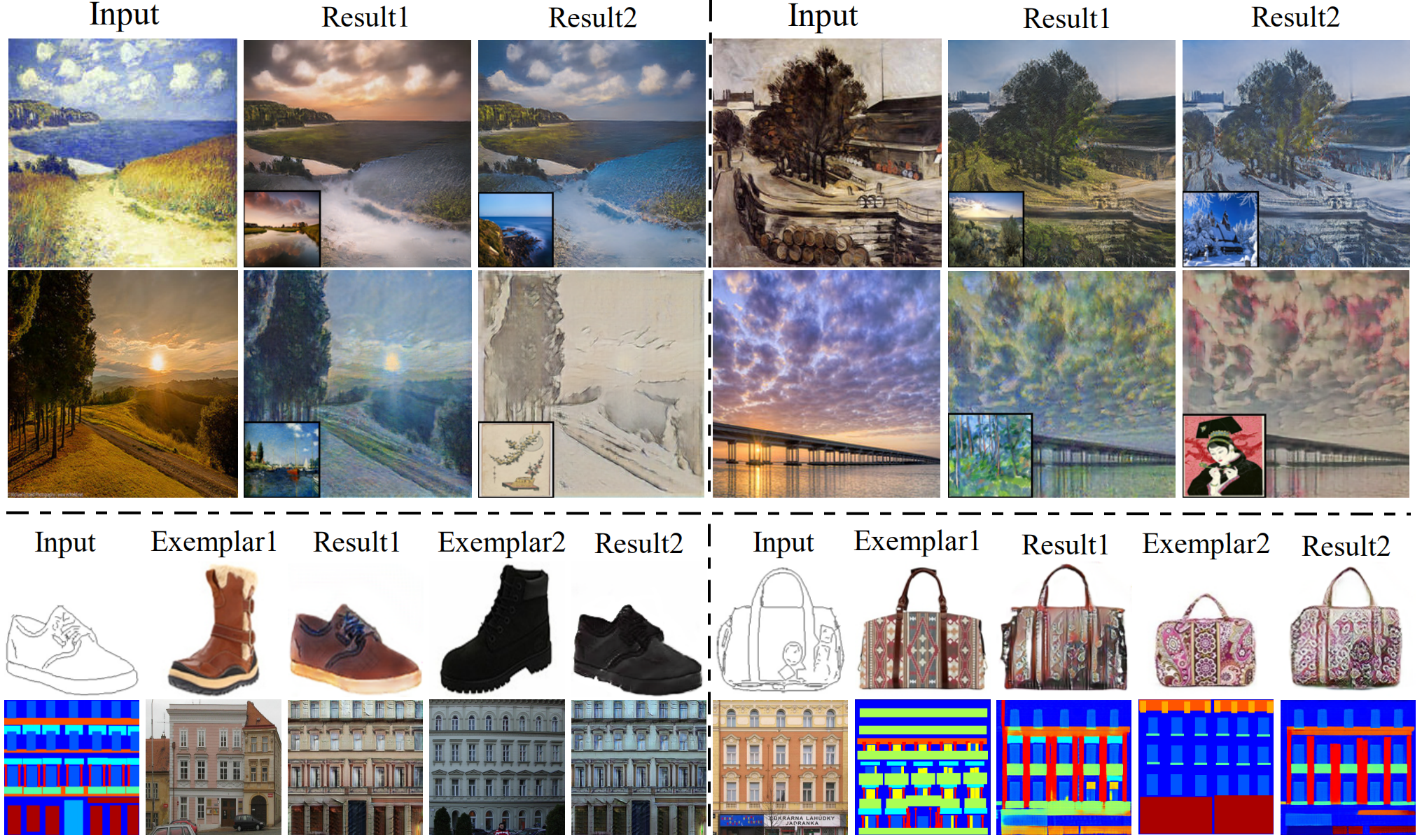}
	\caption{More visual results by our proposed EDIT.} 
	\label{fig:our}
\end{figure*}

\noindent\textbf{Style loss.} For allowing users to control the style by giving an exemplar, a measurement for style difference is required. As advocated in \cite{li2017demystifying}, the batch normalization statistics based loss is adopted instead of the Gram matrix based one, for ease of computation. By denoting $I_{\hat{x}}\defeq\mc{G}(I_x,I_y\in\mb{Y})$ and $I_{\hat{y}}\defeq\mc{G}(I_y,I_x\in\mb{X})$, we have:
\begin{equation}
\begin{aligned}
\zeta_{sty}\defeq&\sum_{l=1}^{N_L}\sum_{m=1}^{M_l}\frac{\bigg((\mu_{\hat{y}}^{l,m}-\mu_{x}^{l,m})^2+(\sigma_{\hat{y}}^{l,m}-\sigma_{x}^{l,m})^2\bigg)}{N_L\times M_l}\\
+&\sum_{l=1}^{N_L}\sum_{m=1}^{M_l}\frac{\bigg((\mu_{\hat{x}}^{l,m}-\mu_{y}^{l,m})^2+(\sigma_{\hat{x}}^{l,m}-\sigma_{y}^{l,m})^2\bigg)}{N_L\times M_l},
\end{aligned}
\label{sty}
\end{equation}
where $N_L$ and $M_l$ are the number of involved layers (in this work, we use the $relu1\_2$, $relu2\_2$, $relu3\_3$, $relu4\_3$ and $relu5\_1$ layers in the VGG16) and that of feature maps in the $l$-th layer. In addition,  $\mu$ and $\sigma$ are the mean and the standard deviation of the corresponding feature map.


\noindent\textbf{Adversarial loss.} The adversarial loss is standard \cite{goodfellow2014generative} as:
\begin{equation}
\begin{aligned}
\zeta_{adv}\defeq& \log \mc{D}(I_x, \mb{X}) +\log (1-\mc{D}(\mc{G}(I_x, I_y\in\mb{Y}),\mb{Y}))\\
+&\log \mc{D}(I_y,\mb{Y})+\log (1-\mc{D}(\mc{G}(I_y, I_x\in\mb{X}),\mb{X})).
\end{aligned}
\label{adv}
\end{equation}

\noindent\textbf{Final objective.} Our optimization is carried out on the total loss, \textit{i.e.} the sum of the above losses, as follows:
\begin{equation}
\begin{aligned}
&\min_{\mc{G}} \max_{\mc{D}}~~ \mathbb{E}_{I_x\sim P_{data}(\mb{X})} \mathbb{E}_{I_y\sim P_{data}(\mb{Y})}~~\zeta_{total},\\
&\text{where}~~ \zeta_{total}\defeq\zeta_{adv}+ \lambda \zeta_{cyc}+ \eta\zeta_{sty},
\end{aligned}
\label{total}
\end{equation}
where $\eta$ and $\lambda$ are coefficients to balance the loss terms. In order to keep the common features effectively, we set $\lambda$  to a relatively large value $10$. As for $\eta$, we observe that setting it in the range from $0.01$ to $0.1$ works well.

\section{Experimental Validation}

\noindent\textbf{Implementation details.}
Our EDIT is implemented in PyTorch and performed on a GeForce RTX 2080 GPU. We use the strategy proposed in \cite{shrivastava2017learning} to improve the training stability, which uses historic generated images to update the parameters of the discriminator. Our optimization adopts an Adam solver. The decays of the first and second order momentums are as default. The learning rate is set to 0.001 at the beginning and linearly decreased as the number of epochs grows. During the training phase, the input images are resized to $256 \times 256$ and augmented by random horizontal flip.\\ 


\noindent\textbf{Competitors \& Evaluation metrics.}
The competitors involved in comparisons contain neural style transfer (NST) \cite{gatys2016image}, cycleGAN \cite{zhu2017unpaired}, metaNST \cite{shen2017meta}, DRIT \cite{lee2018diverse}, MUNIT \cite{huang2018multimodal}, WCT \cite{WCT}, EGSC-IT \cite{SCST}, and art2real \cite{Tomei_2019_CVPR}. The codes of the compared methods are all downloaded from the authors' websites. The elapsed time of testing, model size, and inception score \cite{Salimans2016Improved} are employed as our metrics to quantitatively reveal the performance difference between our EDIT with the other competitors. In addition, to measure how well the content and style are preserved, by following \cite{gatys2016image}, the content error and style error are also adopted. Take the mapping: $\mb{X} \rightarrow \mb{Y}$ as an example, the content error $\mc{E}_{cont}$ is defined as the L2 distance between feature maps of the input image $I_x$ and the generated one $I_x^{y\in\mathbb{Y}}$, \textit{i.e.}
$||\phi_{l}(I_x) - \phi_{l}(I_x^{y\in\mathbb{Y}})||_2$,
where $\phi_{l}(\cdot)$ means the feature maps of the $l$-th layer in the VGG-16 model. The style error is the average L2 distance between the Gram matrices of the generated image $I_x^{y\in\mathbb{Y}}$ and the exemplar  $I_y$. Assume $\text{Gram}_l(\cdot)$, $H_l$, and $W_l$ are the Gram matrix, height and width of the each feature map in the $l$-th layer,  the style error can be expressed as
$\frac{1}{N_L}\sum_{l=1}^{N_L}\frac{1}{4M_l^2H_l^2W_l^2}{||\text{Gram}_l(I_y)-\text{Gram}_l(I_x^{y\in\mathbb{Y}})||_2^2}$. \\

\begin{figure}[htb]
	\centering
	\includegraphics[scale=0.3]{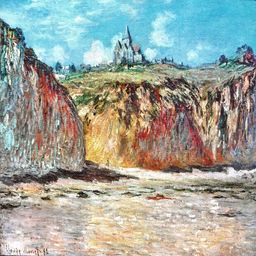}
	\includegraphics[scale=0.3]{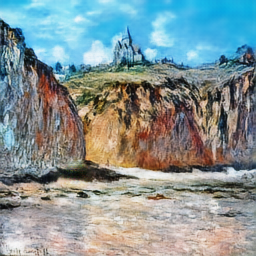}
	\includegraphics[scale=0.3]{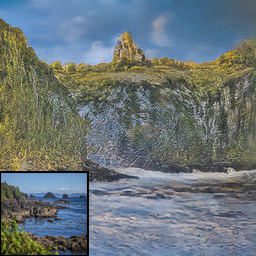}
	\quad \quad \quad Input \quad \quad \quad \quad \quad  art2real \quad \quad \quad \quad \quad \quad Ours
	\caption{Visual comparison between art2real and EDIT}\label{fig:a2p}
\end{figure}

\noindent\textbf{Comparisons.} To quantitatively measure the performance of different competitors, we conduct the experiments on the translation from photo to painiting (Monet). The training and testing data are from \cite{zhu2017unpaired}. The competitors are well-trained on the training data, and tested on the 750 testing data of the photo set with 10 exemplars from the Monet set. This is to say, each compared model generates 7,500 images, on which the content error, the style error, and the inception score are computed.

From Table \ref{tab:comp}, we can observe that in terms of content error, our EDIT slightly falls behind cycleGAN, while outperforms the others. Analogous analysis serves the inception score term. We notice that the I2I translation is a trade-off between the content and the style consistency. Notice that cycleGAN pays more attention on the content loss while only guaranteeing the domain style. As for the style loss, EDIT takes the first place among all the compared methods with a large margin. The cycleGAN is unable to take exemplars as reference, thus we do not provide its style error.
In terms of model size, we provide two sets of comparison: one for one pair domain translation and the other for $n=4$ pairs. Most of the methods including NST, MUNIT, cycleGAN, DRIT, and EGSC-IT require to train multiple models to handle multiple pairs of domain. While EDIT, metaNST and WCT can deal with multiple domain pairs in one model, thanks to either the dynamic parameter generators according to exemplars or the feature transfer manner. Different from metaNST and WCT, EDIT further takes into account the domain knowledge and the cycle consistency. The dynamic parameter generators for both EDIT and metaNST have several fully connected layers, making their models relatively large. A large part of the parameters in metaNST ($10Mb$ vs. $64K$ for shared part) are from the parameter generator, leading to a $870Mb$ storage. Ours demands the parameter network to produce about $3.6Mb$ (vs. $8Mb$ for shared part) parameters dynamically, significantly decreasing the storage ($476Mb$) compared with metaNST. The above verifies, as previously stated in our principle, that the feature extraction can be done using a uniform extractor and part of feature reassembling can also be in common for different images from different domains. In this work, we consider 4 domain pairs, but it is possible to embrace more in the current models of metaNST and EDIT.

\begin{figure*}[t]
	\centering
	\includegraphics[width=1\linewidth]{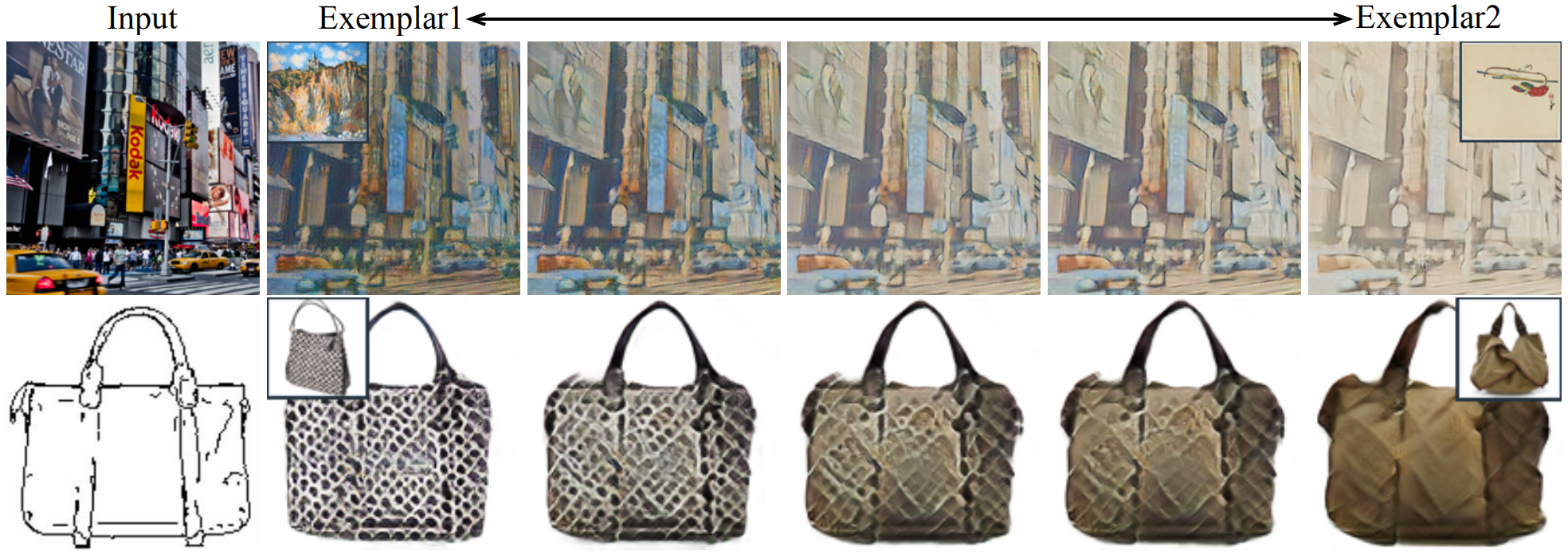}
	\caption{Interpolation results. The left-most column contains two inputs. The second and right-most columns are the results with respect to two different exemplars. The three columns in the middle are the interpolated results by EDIT.}
	\label{fig:inter}
\end{figure*}

In terms of speed, NST takes much longer time, \textit{i.e.} about $420s$ to process a case with size of $256\times 256$, than the others, due to its processing way. The fastest method goes to cycleGAN ($3.5ms$), as it does not need to consider exemplars.  Among the methods that consider exemplars, our EDIT is the most efficient one ($4.3ms$), slightly slower than cycleGAN. In addition, WCT is relatively slower  ($1.7s$) because of the requirement of  SVD operation that has to be executed in CPU. In Table \ref{tab:comp}, we also report the numbers corresponding to EDIT with the discriminator disabled, which reveal the importance of the adversarial mechanism for the target task. Figure \ref{fig:vcomp} depicts three visual comparisons to qualitatively show the difference among the competitors. From the pictures, we can see that our EDIT can very well preserve the content of input and transfer the style of exemplar, making the final results visually striking. It is worth noting that art2real is specifically designed for translation from arts/paintings to realistic photos without using any exemplar, which if reasonably modified, needs multiple models for different domain pairs in nature. Figure \ref{fig:a2p} additionally gives a comparison between art2real and EDIT. The result by art2real indeed has some features of painting removed, however the unnatural-looking of which is still obvious. While, by taking an exemplar into consideration, EDIT produces a more realistic result. We provide other visual results  by EDIT on painting$\leftrightarrow$ painting, edge $\rightarrow$ shoe, edge $\rightarrow$ handbag, and semantic map $\leftrightarrow$ facade in Figure \ref{fig:our} and at our website\renewcommand{\thefootnote}{\fnsymbol{footnote}}\footnote{\url{https://forawardstar.github.io/EDIT-Project-Page/}\label{web}}. Comprehensively, the proposed EDIT is arguably the best candidate. \\

\noindent\textbf{Style interpolation.} One may want to take two or more exemplars/domains as style reference, and produce results simultaneously containing those styles in a controllable fashion. Considering that the dynamic parameters correspond to the exemplars, they can be viewed as their representations in the implicit manifold. Suppose the manifold is continuous and smooth, we can linearly combine the generated parameters to achieve the style interpolation. Figure \ref{fig:inter} displays two cases of style interpolation. The second and the sixth columns offer the translated results by fully using different exemplars. The pictures shown in the middle columns are results by linearly interpolating the parameters of the second and the last columns. As can be seen, via controlling the parameter combination, the visual results vary smoothly between two styles with the content well-preserved.

\section{Concluding Remarks}

In this paper, we have proposed  a network, called EDIT, to translate images from different domains with consideration of specific exemplars in a unified model. The generator of EDIT is built upon a part of blocks configured by shared parameters to uniformly extract features for images from multiple domains, and another part by dynamic parameters exported by an exemplar-domain aware parameter network to catch specific style information. The concepts of cycle consistency and adversarial mechanism make the translation preserve the content and satisfy the distribution of target domain.  We have conducted the extensive experiments to evaluate the performance of EDIT both quantitatively and qualitatively, which reveal the efficacy of our design, and its superiority over the state-of-the-art alternatives. Code is available at \url{https://github.com/ForawardStar/EDIT}.

{\small
\bibliographystyle{ieee}
\bibliography{ref_paper}
}

\end{document}